\documentclass[runningheads]{llncs}

\usepackage{tikz}
\usepackage{comment}
\usepackage{color}

\usepackage{times}
\usepackage{epsfig}
\usepackage{graphicx}
\usepackage{amsmath}
\usepackage{amssymb}
\usepackage{booktabs}
\usepackage{xcolor}
\usepackage{enumitem}
\usepackage{lipsum}
\usepackage{wrapfig}

\usepackage[accsupp]{axessibility}  

\usepackage[width=122mm,left=12mm,paperwidth=146mm,height=193mm,top=12mm,paperheight=217mm]{geometry}

\definecolor{citecolor}{HTML}{0071bc}

\newcommand{\app}{\raise.17ex\hbox{$\scriptstyle\sim$}}

\definecolor{codegreen}{rgb}{0.0,0.6,0.0}

\newcommand{\myparagraph}[1]{{\vspace{.5em} \noindent \bf #1}}

\usepackage[pagebackref,breaklinks,colorlinks]{hyperref}

\usepackage[capitalize]{cleveref}
\crefname{section}{Sec.}{Secs.}
\Crefname{section}{Section}{Sections}
\Crefname{table}{Table}{Tables}
\crefname{table}{Tab.}{Tabs.}

\newcommand{\etal}{\textit{et al}. }

\newcommand{\eg}{\textit{e}.\textit{g}. }

\usepackage[ruled,vlined,linesnumbered]{algorithm2e}
\makeatletter
\newcommand{\algorithmfootnote}[2][\footnotesize]{%
  \let\old@algocf@finish\@algocf@finish
  \def\@algocf@finish{\old@algocf@finish
    \leavevmode\rlap{\begin{minipage}{\linewidth}
    #1#2
    \end{minipage}}%
  }%
}
\makeatother

\begin{document}
\pagestyle{headings}
\mainmatter
\def\ECCVSubNumber{317}  

\title{Robust Multi-Object Tracking by Marginal Inference} 

\titlerunning{Robust Multi-Object Tracking by Marginal Inference}
%
\author{Yifu Zhang\inst{1\dag} \and
Chunyu Wang\inst{2} \and
Xinggang Wang\inst{1} \and
Wenjun Zeng\inst{3} \and
Wenyu Liu\inst{1\ddag}}
\authorrunning{Y. Zhang et al.}
%
\institute{Huazhong University of Science and Technology \and
Microsoft Research Asia \and Eastern Institute for Advanced Study
}
\maketitle

\vspace{-5mm}

\begin{abstract}
Multi-object tracking in videos requires to solve a fundamental problem of one-to-one assignment between objects in adjacent frames. Most methods address the problem by first discarding impossible pairs whose feature distances are larger than a threshold, followed by linking objects using Hungarian algorithm to minimize the overall distance. However, we find that the distribution of the distances computed from Re-ID features may vary significantly for different videos. So there isn't a single optimal threshold which allows us to safely discard impossible pairs. To address the problem, we present an efficient approach to compute a marginal probability for each pair of objects in real time. The marginal probability can be regarded as a normalized distance which is significantly more stable than the original feature distance. As a result, we can use a single threshold for all videos. The approach is general and can be applied to the existing trackers to obtain about one point improvement in terms of IDF1 metric. It achieves competitive results on MOT17 and MOT20 benchmarks. In addition, the computed probability is more interpretable which facilitates subsequent post-processing operations.
\end{abstract}

\let\thefootnote\relax\footnotetext{$^{\dag}$ This work was done when Yifu Zhang was an intern of Microsoft Research Asia. $^{\ddag}$ Corresponding author.}

\vspace{-5mm}
\section{Introduction}

Multi-object tracking (MOT) is one of the most active topics in computer vision. The state-of-the-art methods \cite{wojke2017simple,wang2019towards,zhang2021fairmot,zhou2020tracking,lu2020retinatrack,pang2020quasi,shan2020fgagt,sun2020transtrack} usually address the problem by first detecting objects in each frame, and then linking them to trajectories based on Re-ID features. Specifically, it computes distances between objects in adjacent frames, discards impossible pairs with large distances, and determines the matched pairs by minimizing the overall distance by applying the Hungarian algorithm \cite{kuhn1955hungarian}. 

\begin{figure}[!htbp]
	\centering
	\includegraphics[width=0.75\linewidth]{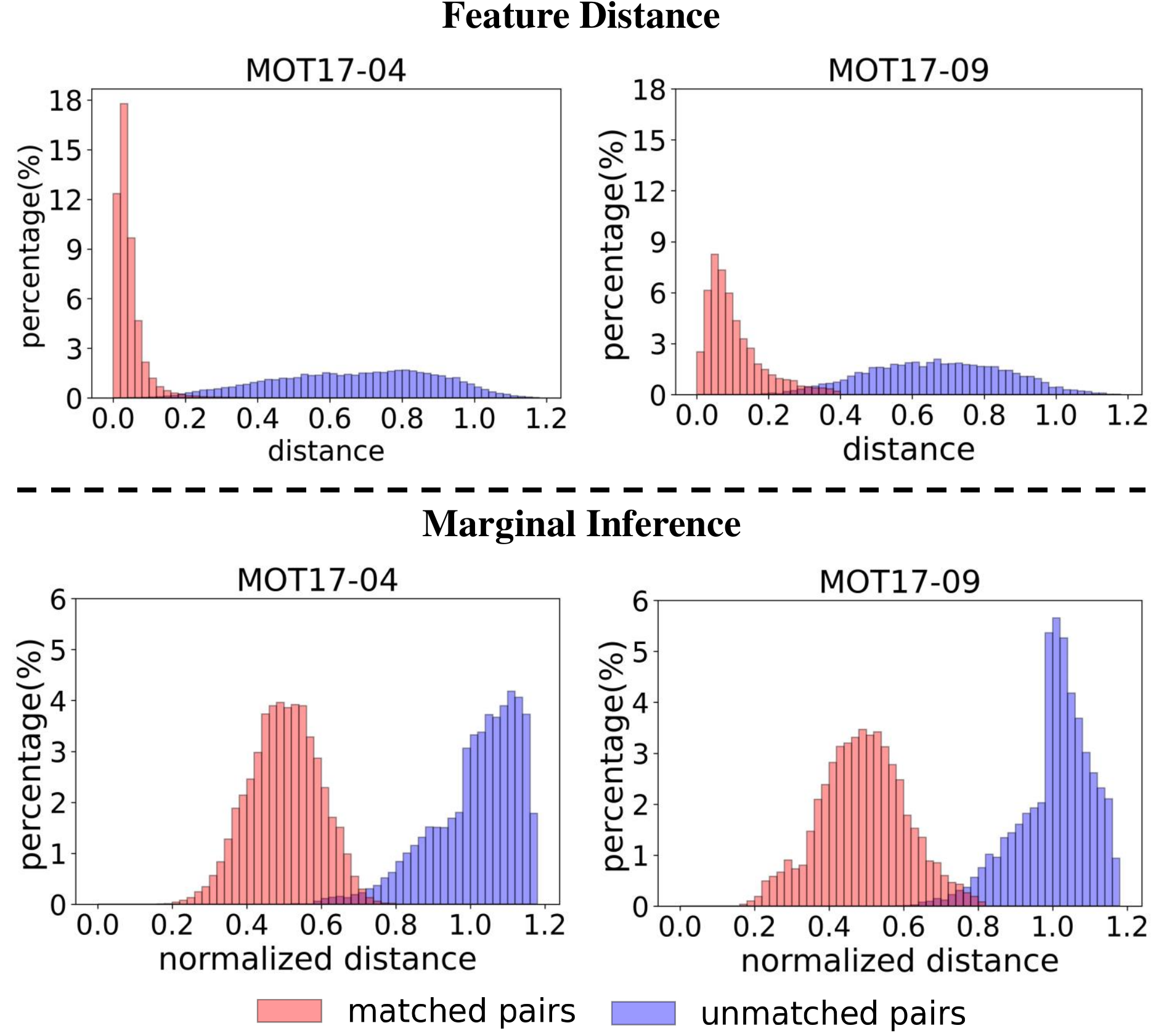}
	\caption{Distance distribution of the matched pairs and unmatched pairs, respectively, on two videos. The top shows the distances directly computed from Re-ID features the bottom shows our normalized distances (marginal probability).}
	\label{fig:teasing}
	\vspace{-3mm}
\end{figure}

The core of the linking step is to find a threshold where the distances between the matched objects are smaller than it, while those of the unmatched ones are larger than it. The threshold setting is done by experience and has not received sufficient attention. However, our experiment shows that even the best Re-ID model cannot discard all impossible pairs without introducing false negatives using a single threshold because the distances may vary significantly on different frames as shown in Figure \ref{fig:teasing} (top). We can see that the optimal threshold to discriminate matched and unmatched pairs is $0.2$ for video ``MOT17-04'' and $0.4$ for ``MOT17-09'' which are very different.

We argue in this work that we should put a particular value of distance into context when we determine whether it is sufficiently small to be a matched pair. For example, in Figure \ref{fig:teasing} (top), $0.2$ is a large distance for video ``MOT17-04'' but it is a small one for ``MOT17-09'' considering their particular distance distributions. To achieve this goal, we propose to compute a marginal probability for each pair of objects being matched by considering the whole data association space which consists of all possible one-to-one assignment structures. The marginal probability is robust to distance distribution shift and significantly improves the linking accuracy in our experiment. For example, in the ``Public Detection'' track of the MOT17 challenge, the IDF1 score improves from $59.6\%$ to $65.0\%$.

We consider a possible matching between all the detections and trajectories as one structure. However, naively enumerating all structures is intractable especially when the number of objects in videos is large. We address the complexity issue by computing a small number of low-cost supporting structures which often overlap with the maximum a posterior solution found by Hungarian algorithm \cite{bewley2016simple,wojke2017simple}. The marginal probability of each pair is computed by performing marginal inference among these structures. Our experiments on videos with a large number of objects show that it takes only a fraction of the time to compute without affecting the inference speed.

The approach is general and applies to almost all existing multi-object trackers. We extensively evaluate approach with the state-of-the-art trackers on multiple datasets. It consistently improves the tracking performances of all methods on all datasets with little extra computation. In particular, we empirically find that it is more robust to occlusion. When occlusion occurs, the distance between two (occluded) instances of the same person becomes larger and the conventional methods may treat them as two different persons. However, the marginal probability is less affected and the two instances can be correctly linked.




\section{Related work}
Most \textit{state-of-the-art} multi-object tracking methods \cite{wojke2017simple,wang2019towards,zhang2021fairmot,zhou2020tracking,lu2020retinatrack,pang2020quasi,shan2020fgagt,zhang2021bytetrack,unicorn} follow the \textit{tracking-by-detection} paradigm which form trajectories by
associating detections in time. They first adopt detectors such as \cite{ren2015faster,redmon2018yolov3,zhou2019objects,lin2017focal,fang2022unleashing} to get the location of the objects and then link the detections to the existing trajectories according to similarity. Similarity computation and matching strategy are two key components of data association in the \textit{tracking-by-detection} paradigm. We review different methods from the two aspects and compare them to our approach.

\subsection{Similarity computation}
Location, motion, and appearance are three important cues to compute similarity between detections and tracks. IOU-Tracker \cite{bochinski2017high} computes the spatial overlap of detections in neighboring frames as similarity. SORT \cite{bewley2016simple} adopts Kalman Filter \cite{welch1995introduction} as a motion model to predict the future locations of objects. The similarity is computed by the IoU of predicted locations and detected locations. The two trackers are widely used in practice due to their speed and simplicity. However, both the two trackers will cause a large number of identity switches when encountering camera motion and crowded scenes. To decrease the identity switches, DeepSORT \cite{wojke2017simple} adopts a deep neural network to extract appearance features as appearance cues can refind lost objects. The final similarity is a weighted sum of motion similarity computed by Kalman Filter and cosine similarity of the appearance features. Bae \etal \cite{bae2014robust} also proposes an online discriminative appearance learning method to handle similar appearances of different objects. Many \textit{state-of-the-art} methods \cite{chen2018real,voigtlaender2019mots,wang2019towards,zhang2021fairmot,lu2020retinatrack,pang2020quasi,zhang2022voxeltrack} follow \cite{wojke2017simple} to compute the similarity using location, motion and appearance cues. Some methods \cite{xu2019spatial,shan2020fgagt} utilize networks to encode appearance and location cues into similarity score. Recently, some methods \cite{wang2019towards,zhang2021fairmot,lu2020retinatrack,pang2020quasi} combine the detection task and re-identification task in a single neural network to reduce computation cost. JDE \cite{wang2019towards} first proposes a joint detection and embedding model to first achieve a (near) real-time MOT system. FairMOT \cite{zhang2021fairmot} deeply studies the reasons for the unfairness between detection and re-identification task in anchor-based models and proposes a high-resolution anchor-free model to extract more discriminative appearance features. QDTrack \cite{pang2020quasi} densely samples hundreds of region proposals on a pair of images for contrastive learning of appearance features to make use of the majority of the informative regions on the images. It gets very high-quality appearance features and can achieve \textit{state-of-the-art} results only using appearance cues.

Our method also uses the location, motion, and appearance cues to compute similarity. However, we find that the appearance feature distance distribution may vary significantly for different videos. We achieve a more stable distribution by computing the marginal probability based on appearance features.

\subsection{Matching strategy}
After computing similarity, most methods \cite{bewley2016simple,wojke2017simple,bae2014robust,chen2018real,voigtlaender2019mots,wang2019towards,zhang2021fairmot,xu2019spatial,shan2020fgagt} use Hungarian Algorithm \cite{kuhn1955hungarian} to complete matching. Bae \etal \cite{bae2014robust} matches tracklets in different ways according to their confidence values. Confident tracklets are locally matched with online-provided detections and fragmented tracklets are globally matched with confident tracklets or unmatched detections. The advantage of confidence-based tracklets matching is that it can handle track fragments due to occlusion or unreliable detections. DeepSORT \cite{wojke2017simple} proposes a cascade matching strategy, which first matches the most recent tracklets to the detections and then matches the lost tracklets. This is because recent tracklets are more reliable than lost tracklets. MOTDT \cite{chen2018real} proposes a hierarchical matching strategy. It first associates using the appearance and motion cues. For the unmatched tracklets and detections (usually under severe occlusion), it matches again by IoU. JDE \cite{wang2019towards} and FairMOT \cite{zhang2021fairmot} also follow the hierarchical matching strategy proposed by MOTDT. QDTrack \cite{pang2020quasi} applies a bi-directional softmax to the appearance feature similarity and associates objects with a simple nearest neighbor search. All these methods need to set a threshold to decide whether the detections match the tracklets. If the distance is larger than the threshold, the matching is rejected. It is very challenging to set an optimal threshold for all videos because the distance distribution may vary significantly as the appearance model is data-driven. The work most related to our approach is Rezatofighi \etal \cite{rezatofighi2015joint} which also uses probability for matching. However, their motivation and the solution to compute probability are different from ours. 

We follow the matching strategy of \cite{chen2018real,wang2019towards,zhang2021fairmot} which hierarchically uses appearance, motion, and location cues. The main difference is that we turn the appearance similarity into marginal probability (normalized distance) to achieve a more stable distance distribution. The motion model and location cues are more generalized, so we do not turn them into probability. The matching is also completed by the Hungarian Algorithm. Our marginal probability is also more robust to occlusion as it decreases the probability of false matching. We can thus set a looser (higher) matching threshold to refind some lost/occluded objects.

\section{Method}
\subsection{Problem formulation}
Suppose we have $M$ detections and $N$ history tracks at frame $t$, our goal is to assign each detection to one of the tracks which has the same identity. Let $\textbf{d}^1_t,...,\textbf{d}^M_t$ and $\textbf{h}^1_t,...,\textbf{h}^N_t$ be the Re-ID features of all $M$ detections and $N$ tracks at frame $t$, respectively. We compute a cosine similarity matrix $\textbf{S}_t \in [0,1]^{M\times N}$ between all the detections and tracks as follows:
\begin{equation}
\label{eq:sim}
    \textbf{S}_t(i,j) = \frac{\textbf{d}_t^i \cdot \textbf{h}_t^j}{||\textbf{d}_t^i|| \cdot ||\textbf{h}_t^j||},
\end{equation}
where $i \in \{1,...,M\}$ and $j \in \{1,...,N\}$. For simplicity, we replace $\{1,...,M\}$ by $\mathbb{M}$ and $\{1,...,N\}$ by $\mathbb{N}$ in the following myparagraphs.

Based on the similarity $\textbf{S}_t$, we compute a marginal probability matrix $\textbf{P}_t \in [0,1]^{M\times N}$ for all pairs of detections and tracks. $\textbf{P}_t(i,j)$ represents the marginal probability that the $i_{th}$ detection is matched to the $j_{th}$ track. We compute $\textbf{P}_t(i,j)$ considering all possible matchings. Let $\mathbb{A}$ denote the space which consists of all possible associations (or matchings). Under the setting of multi-object tracking, each detection matches at most one track and each track matches at most one detection. We define the space $\mathbb{A}$ as follow:
\begin{align}
\label{eq:space}
    \mathbb{A} &= \Big \{A = \Big (m_{ij}\Big )_{i\in \mathbb{M},j\in \mathbb{N}} \; \Big | \; m_{ij} \in \{0,1\} \\
               & \land \; \sum_{i=0}^{M} m_{ij} \leqslant 1, \forall j \in \mathbb{N} \\
               & \land \; \sum_{j=0}^{N} m_{ij} \leqslant 1, \forall i \in \mathbb{M} \Big \},
\end{align}
where $A$ is one possible matching. We define $\mathbb{A}_{ij}$ as a subset of $\mathbb{A}$, which contains all the matchings where the $i_{th}$ detection is matched to the $j_{th}$ track:
\begin{equation}
    \mathbb{A}_{ij} = \{A \in \mathbb{A} \; | \; m_{ij} = 1\}
\end{equation}
The marginal probability $\textbf{P}_t(i,j)$ can be computed by marginalizing $\mathbb{A}_{ij}$ as follows:
\begin{equation}
\label{eq:marginal}
    \textbf{P}_t(i,j) = \sum_{A \in \mathbb{A}_{ij}}p(A), 
\end{equation}
where $p(A)$ is a joint probability representing the probability of one possible matching $A$ and can be computed as follows:
\begin{equation}
    p(A) = \prod_{\forall q\in \mathbb{M}, \forall r\in \mathbb{N}} \Bigg(\frac{\exp \big (\textbf{S}_t(q,r) \big)}{\sum\nolimits_{r=1}^{N} \exp \big (\textbf{S}_t(q,r) \big)})\Bigg)
\end{equation}
The most difficult part to obtain $\textbf{P}_t(i,j)$ is to computing all possible matchings in $\mathbb{A}_{ij}$ because the total number of matchings is \textit{n}-permutation.

\subsection{Our solution}
We consider the structured problems. We view each possible matching between all the detections and tracks as a structure $\textbf{k} \in \{0,1\}^{MN}$, which can be seen as a flattening of the matching matrix. We define all the structures in the space $\mathbb{A}$ as $\textbf{K} \in \{0,1\}^{MN \times D}$, where $D$ is the number of all possible matchings and $MN \ll D$.

We often use the structured log-potentials to parametrize the structured problems. The scores of the structures can be computed as $\boldsymbol{\theta} := \textbf{K}^\top \boldsymbol{S}$, where $\boldsymbol{S} \in \mathbb{R}^{MN}$ is a flattening of the similarity matrix $\textbf{S}_t$. Suppose we have variables $V$ and factors $F$ in a factor graph \cite{kschischang2001factor}, $\boldsymbol{\theta}$ can be computed as:
\begin{equation}
    \theta_o := \sum_{v \in V}S_{V,v}(o_v) + \sum_{f \in F} S_{F,f}(o_f),
\end{equation}
where $o_v$ and $o_f$ are local structures at variable and factor nodes. $\boldsymbol{S}_V$ and $\boldsymbol{S}_F$ represent the log-potentials. In our linear assignment setting, we only have variables and thus $\boldsymbol{\theta}$ can be written in matrix notation as $\boldsymbol{\theta} = \textbf{K}^{\top} \boldsymbol{S}_V$. 

The optimal matching between detections and tracks can be viewed as the MAP inference problem, which seeks the highest-scoring structure. It can be rewritten using the structured log-potentials as follows:
\begin{align}
    \text{MAP}_\textbf{K}(\boldsymbol{S}) &:= \mathop{\arg\max}\limits_{\textbf{v}:=\textbf{K}\textbf{y},\textbf{y} \in \triangle^{D}} \boldsymbol{\theta}^{\top} \textbf{y} \\
    & \; = \mathop{\arg\max}\limits_{\textbf{v}:=\textbf{K}\textbf{y},\textbf{y} \in \triangle^{D}} {\boldsymbol{S}_V}^{\top} \textbf{v}, 
\end{align}
where $\textbf{v} \in \{0,1\}^{MN}$ is the highest-scoring structure and $\{\textbf{v} = \textbf{K}\textbf{y}, \textbf{y} \in \triangle^D\}$ is the Birkhoff polytope \cite{birkhoff1946tres}. In linear assignment, the structure $\textbf{v}$ can be obtained by Hungarian algorithm \cite{kuhn1955hungarian}. 

The main challenge of computing the marginal probability as in Equation~\ref{eq:marginal} is that the total number of structures $D$ is very large and usually not tractable. To address the problem, we propose to compute a small number of low-cost and often-overlapping structures instead of enumerating all of them. In \cite{niculae2018sparsemap}, Niculae \etal show that this can be achieved by regularizing the MAP inference problem with a squared $l_2$ penalty on the returned posteriors which was inspired by \cite{martins2016softmax}. Computing multiple structures which approximate the MAP inference problem can be written as follows:
\begin{align}
\label{eq:sparsemap}
    \text{L2MAP}_\textbf{K}(\boldsymbol{S}) &:= \mathop{\arg\max}\limits_{\textbf{v}:=\textbf{K}\textbf{y},\textbf{y} \in \triangle^{D}} \boldsymbol{\theta}^{\top} \textbf{y} - \frac{1}{2}||\textbf{K}\textbf{y}||^2_2\\
    & \; = \mathop{\arg\max}\limits_{\textbf{v}:=\textbf{K}\textbf{y},\textbf{y} \in \triangle^{D}} {\boldsymbol{S}_V}^{\top} \textbf{v} - \frac{1}{2}||\textbf{v}||^2_2, 
\end{align}
The result is a quadratic optimization problem and it can be solved by the conditional gradient (CG) algorithm \cite{lacoste2015global}. The Equation~\ref{eq:sparsemap} can be written by function $f$ as follows:
\begin{equation}
    f(\textbf{v}) := {\boldsymbol{S}_V}^{\top} \textbf{v} - \frac{1}{2}||\textbf{v}||^2_2,
\end{equation}
A linear approximation to $f$ around a point $\textbf{v}^{\prime}$ is:
\begin{equation}
    \hat{f}(\textbf{v}) := (\nabla_{\textbf{v}}f)^{\top}\textbf{v} = (\boldsymbol{S}_V - \textbf{v}^{{\prime}})^{\top}\textbf{v},
\end{equation}
We can turn the optimization problem of $\hat{f}$ into an MAP inference problem. The variable scores of the MAP inference problem at each step is $\boldsymbol{S}_V - \textbf{v}^{{\prime}}$. At each step, we use Hungarian algorithm to solve the MAP inference problem and get a high-scoring structure $\textbf{z} \in \{0,1\}^{MN}$. Then we use $\textbf{z}$ to substitute $\textbf{v}^{{\prime}}$ for another step. After a small number of steps, we obtain a set of high-scoring and often-overlapping structures $\mathbb{Z} = \{\textbf{z}_1,...,\textbf{z}_n\}$, where $n$ is the number of steps. We compute the marginal probability $\textbf{P}_t$ by marginalizing $\mathbb{Z}$ following Equation~\ref{eq:marginal}:
\begin{equation}
\label{eq:pro}
    \textbf{P}_t(i,j) = \sum_{\textbf{z} \in \mathbb{Z}_{ij}} \Bigg(\frac{\exp (-\textbf{C}_t^{\top}\textbf{z})}{\sum_{v=1}^n \exp (-\textbf{C}_t^{\top} \textbf{z}_v)} \Bigg)
\end{equation}
where $\mathbb{Z}_{ij}$ contains all structures that the $i_{th}$ detection is matched to the $j_{th}$ track and $\textbf{C}_t \in [0,1]^{MN}$ is a flattened feature distance matrix $\boldsymbol{S}_V$.

\subsection{Tracking algorithm}
Our tracking algorithm jointly considers appearance, motion, and location cues. In each frame, we adopt Hungarian algorithm \cite{kuhn1955hungarian} to perform matching two times hierarchically: 1) marginal probability matching, 2) IoU matching. 

We first define some thresholds in our tracking algorithm. Thresholds $C_t$ and $C_d$ are the confidence thresholds for the detections. Thresholds $T_p$ and $T_{IoU}$ are for the marginal probability matching and IoU matching, respectively. 

In the first frame, we initialize all detections with scores larger than $C_d$ as new tracks. In the following frame $t$, we first match between all $N$ tracks and all $M$ medium detections using marginal probability $\textbf{P}_t \in [0,1]^{M \times N}$ calculated by Equation~\ref{eq:pro} and the Mahalanobis distance $\textbf{M}_t \in \mathbb{R}^{M \times N}$ computed by Kalman Filter proposed in \cite{wojke2017simple}. The cost matrix is computed as follows:
\begin{equation}
    \textbf{D}_p = \omega (1 - \textbf{P}_t) + (1 - \omega) \textbf{M}_t, 
\end{equation}
where $\omega$ is a weight that balances the appearance cue and the motion cue. We set $\omega$ to be 0.98 in our experiments. We adopt Hungarian algorithm to perform the first matching and we reject the matching whose distance is larger than $T_p$. It is worth noting that $T_p$ may vary significantly for different frames or videos if we directly utilize the appearance feature similarity for matching. The marginal probability matching has a significantly more stable $T_p$.

For the unmatched detections and tracks, we perform the second matching using IoU distance with the threshold $T_{IoU}$. This works when appearance features are not reliable (\eg occlusion). 

Finally, we mark lost for the unmatched tracks and save it for 30 frames. For the unmatched detections, if the score is larger than $C_t$, we initialize a new track. We also update the appearance features following \cite{zhang2021fairmot}.

\section{Experiments}

\subsection{MOT benchmarks and metrics}
\myparagraph{Datasets. } We evaluate our approach on MOT17 \cite{milan2016mot16} and MOT20 \cite{MOTChallenge20} benchmarks. The two datasets both provide a training set and a test set, respectively. The MOT17 dataset has videos captured by both moving and stationary cameras from various viewpoints at different frame rates. The videos in the MOT20 dataset are captured in very crowded scenes so there is a lot occlusion happening. There are two evaluation protocols which either use the provided public detections or private detections generated by any detectors. In particular, the MOT17 dataset provides three sets of public detections generated DPM \cite{felzenszwalb2009object}, Faster R-CNN \cite{ren2015faster} and SDP \cite{yang2016exploit}, respectively and we evaluate our approach on all of them. The MOT20 dataset provides one set of public detections generated by Faster R-CNN. 

\myparagraph{Metrics. } We use the CLEAR metric \cite{bernardin2008evaluating} and IDF1 \cite{ristani2016performance} to evaluate different aspects of multi-object tracking. Multi-Object Tracking Accuracy (MOTA) and Identity F1 Score (IDF1) are two main metrics. MOTA focuses more on the detection performance. IDF1 focuses on identity preservation and depends more on the tracking performance. 

\subsection{Implementation details}
We evaluate our approach with two existing feature extractors. The first is the state-of-the-art one-stage method FairMOT \cite{zhang2021fairmot} which jointly detects objects and estimates Re-ID features in a single network. The second is the state-of-the-art two-stage method following the framework of DeepSORT \cite{wojke2017simple} which adopts Scaled-YOLOv4 \cite{wang2020scaled} as the detection model and BoT \cite{luo2019bag} as the Re-ID model. The input image is resized to $1088 \times 608$. In the linking step, we set $C_d = 0.4$, $C_t = 0.5$, $T_p = 0.8$, $T_{IoU} = 0.5$. We set the number of steps $n$ to be 100 when computing the marginal probability. The inference speed of the models is listed as follows: 40 FPS for Scaled-YOLOv4, 26 FPS for FairMOT and 17 FPS for BoT. 

\myparagraph{Public detection. } 
In this setting, we adopt FairMOT \cite{zhang2021fairmot} and our marginal inference data association method. Following the previous works of Tracktor \cite{bergmann2019tracking} and CenterTrack \cite{zhou2020tracking}, we only initialize a new trajectory if it is near a public detection (\eg IoU is larger than a threshold). In particular, we set a strict IoU theshold 0.75 to make our trajectories as close to the public detections as possible. The FairMOT model is pre-trained on the COCO dataset \cite{lin2014microsoft} and finetuned on the training set of MOT17 and MOT20, respectively.


\myparagraph{Private detection. }
We adopt the detection model Scaled-YOLOv4 \cite{wang2020scaled} and Re-ID model BoT \cite{luo2019bag} implemented by FastReID \cite{he2020fastreid} in the private detection setting. We train Scaled-YOLOv4 using the YOLOv4-P5 \cite{wang2020scaled} model on the same combination of different datasets as in FairMOT \cite{zhang2021fairmot}. We train BoT on Market1501 \cite{zheng2015scalable}, DukeMTMC \cite{ristani2016performance} and MSMT17 \cite{wei2018person}. All the training process is the same as the references except the training data. For the Re-ID part, we multiply the cosine distance by 500 to make it evenly distributed between 0 and 1. 

\myparagraph{Ablation study. }
For ablation study, we evaluate on the training set of MOT17 and MOT20. To make it more similar to real-world applications, our training data and evaluation data have different data distribution. We also use different detection models and Re-ID models to evaluate the generalization ability of our method. We select three models, FairMOT, Scaled-YOLOv4, and BoT. We adopt FairMOT either as a joint detection and Re-ID model or a separate Re-ID model. We adopt Scaled-YOLOv4 as a detection model and BoT as a Re-ID model. We train Scaled-YOLOv4 on the CrowdHuman \cite{shao2018crowdhuman} dataset. We train FairMOT on the HiEve \cite{lin2020human} dataset. We train BoT on the Market1501, DukeMTMC, and MSMT17 datasets. The matching threshold is 0.4 for the distance-based method and 0.8 for the probability-based method.

\subsection{Evaluation of the marginal probability}

\myparagraph{More stable distribution. }
In this part, we try to prove that marginal probability is more stable than feature distance. We compare the IDF1 score of the two different methods and also plot the distance distribution between detections and tracks of each method. We adopt different detection models and Re-ID models and evaluate on different datasets. The results are shown in Table~\ref{table:dp}. We can see that the marginal probability matching method has about 1 point IDF1 score higher than the distance-based matching method in most settings. The optimal threshold for each video is different. However, it is not realistic to set different thresholds for different videos in some testing scenarios or real-world applications. So we set one threshold for all videos in a dataset. One reason for the performance gain is that the optimal threshold for each video is similar in the probability-based method, which means a single threshold is suitable for most videos. 

To obtain the distance distribution, we adopt Scaled-YOLOv4 \cite{wang2020scaled} as the detector and FairMOT \cite{zhang2021fairmot} as the Re-ID model to perform tracking on the training set of MOT17. We find the optimal threshold for each video by grid search and plot the distance distribution between detections and tracks in each video. The saved distance is the input of the first matching in our tracking algorithm. As is shown in Figure~\ref{fig:teasing}, the marginal probability distribution of each video is similar and is more stable than feature distance distribution. Also, the optimal threshold for each video is similar in the probability-based matching method and varies significantly in the distance-based matching method. 

\begin{table}
\begin{center}
\setlength{\tabcolsep}{10pt}
\begin{tabular}{l|ll|l|ll}
\toprule
Dataset & Det & ReID & Match & MOTA$\uparrow$ & IDF1$\uparrow$\\
\midrule
MOT17 & FairMOT & FairMOT & D & 47.6 & 58.0\\
MOT17 & FairMOT & FairMOT & P & 47.6 & \textbf{58.2 (+0.2)}\\
\midrule
MOT17 & Scaled-YOLOv4 & FairMOT & D & 71.5 & 72.5\\
MOT17 & Scaled-YOLOv4 & FairMOT & P & 71.4 & \textbf{73.6 (+1.1)}\\
\midrule
MOT17 & Scaled-YOLOv4 & BoT & D & 71.5 & 74.8\\
MOT17 & Scaled-YOLOv4 & BoT & P & 71.6 & \textbf{75.7 (+0.9)}\\
\midrule
MOT20 & FairMOT & FairMOT & D & 43.2 & 45.9\\
MOT20 & FairMOT & FairMOT & P & 43.1 & \textbf{46.7 (+0.8)}\\
\midrule
MOT20 & Scaled-YOLOv4 & FairMOT & D & 73.9 & 69.3\\
MOT20 & Scaled-YOLOv4 & FairMOT & P & 74.1 & \textbf{70.1 (+0.8)}\\
\midrule
MOT20 & Scaled-YOLOv4 & BoT & D & 74.0 & 69.5\\
MOT20 & Scaled-YOLOv4 & BoT & P & 74.2 & \textbf{70.2 (+0.7)}\\
\bottomrule
\end{tabular}
\end{center}
\caption{Comparison of probability-based method and distance-based method. ``P'' is short for probability and ``D'' is short for distance. ``Det'' is short for detection model and ``Re-ID'' is short for Re-ID model. }
\vspace{-5mm}
\label{table:dp}
\end{table}

\begin{figure}
	\centering
	\includegraphics[width=3in]{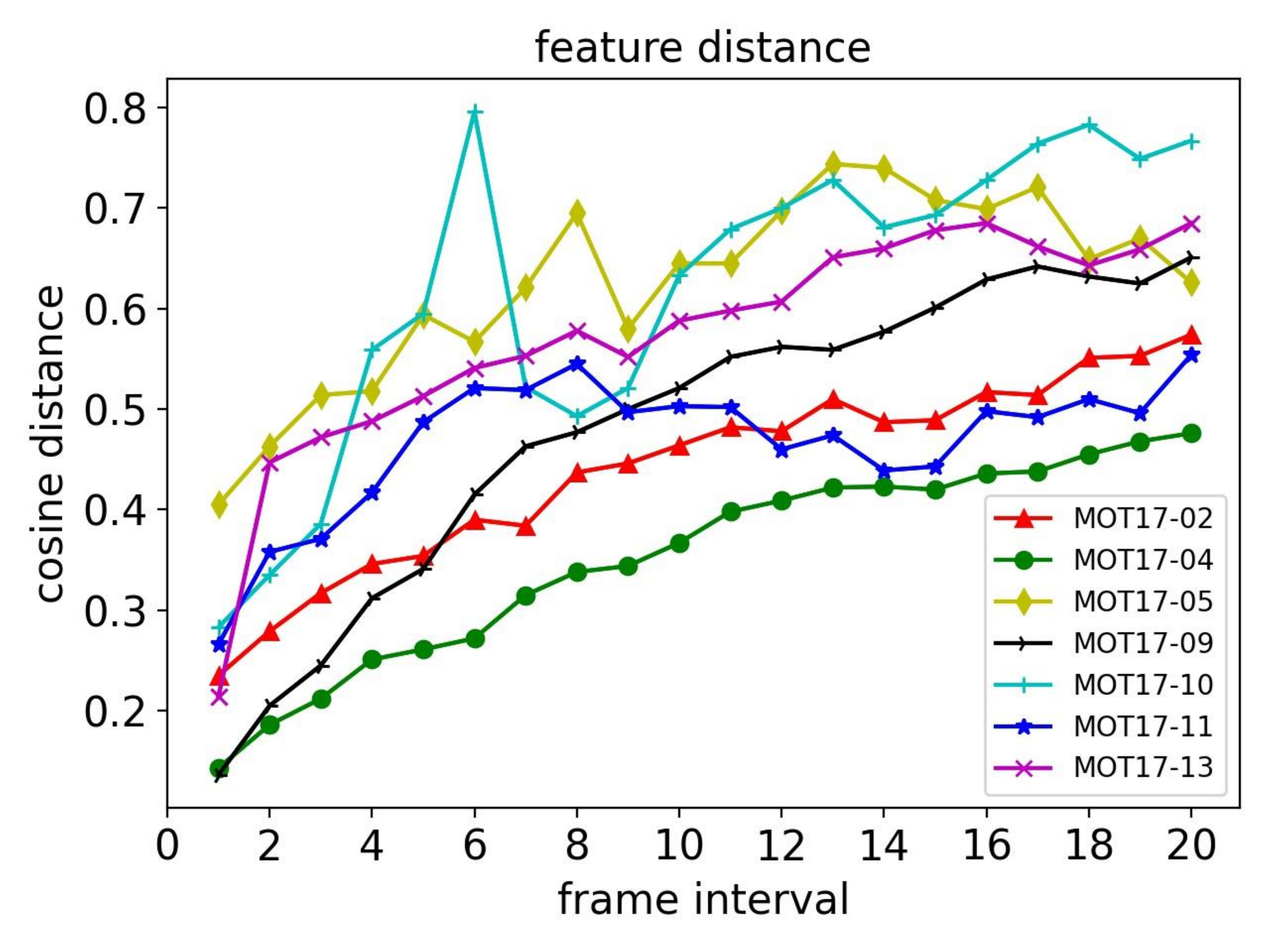}
	\caption{Visualization of cosine distances of the appearance features at different frame intervals. The appearance features are extracted by the BoT Re-ID model. We show the results of all the sequences from the MOT17 training set. 
	}
	\label{fig:interval}
	\vspace{-3mm}
\end{figure}

\myparagraph{More robust to occlusion. }
There are many occlusion cases in multi-object tracking datasets \cite{milan2016mot16,MOTChallenge20}. Even state-of-the-art detectors cannot detect objects under severe occlusions. Therefore, in many cases, an object will reappear after being occluded for a few frames. The key to getting a high IDF1 score is to preserve the identities of these reappeared objects, which is also a main challenge in multi-object tracking. Using appearance features is an effective way to preserve the identities of the occluded objects. However, we find in our experiments that the appearance feature distances of the same object increases linearly with the increase of the number of interval frames and the distance becomes even larger in the case of occlusion, as is shown in Figure~\ref{fig:interval}. The feature distance becomes very large after 20 frames. Thus, it is difficult to retrieve the lost object because it needs a high matching threshold, which may lead to other wrong matchings. 

\begin{figure*}
	\centering
	\includegraphics[width=4.5in]{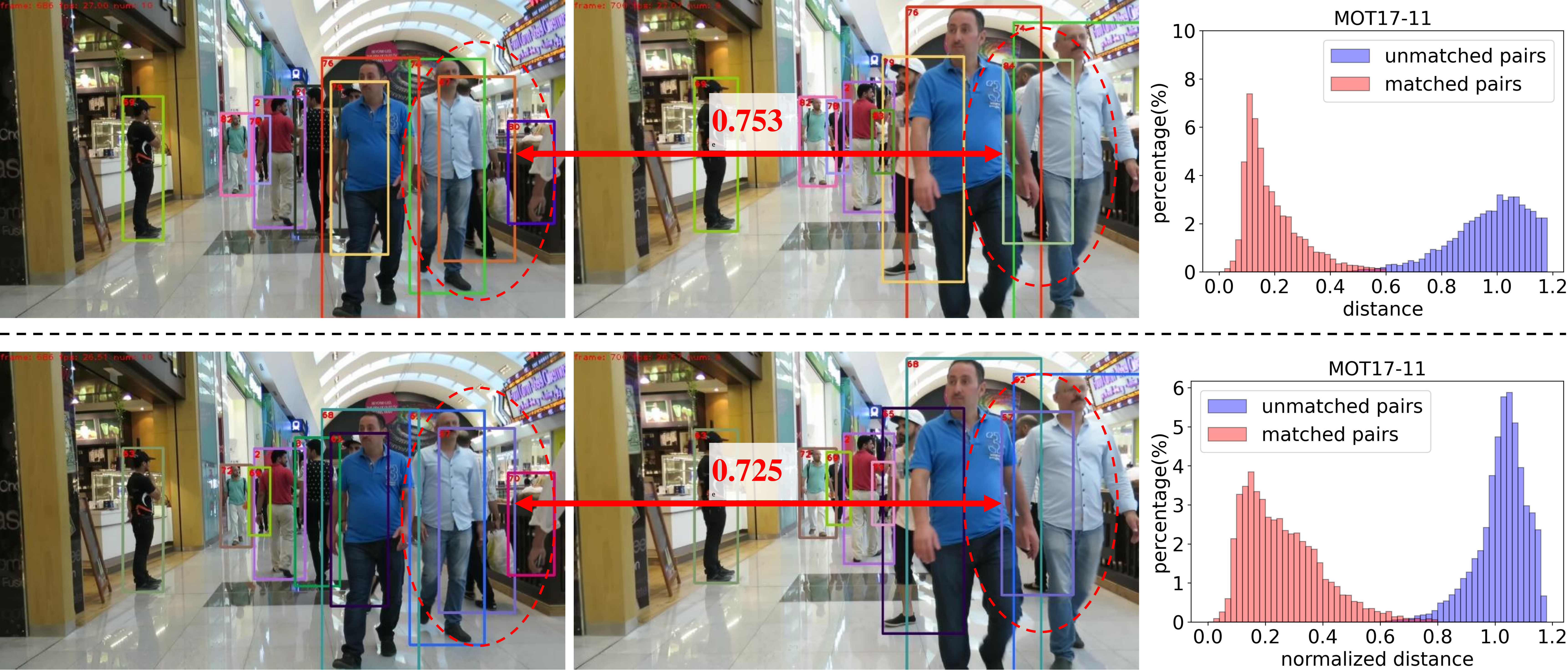}
	\caption{Visualization of how our approach preserves object identity in the case of occlusion. The first line is the distanced-based method and the second line is the probability-base method. We show tracking visualizations, distance values, and distance distribution for both methods. The tracking results are from frame 686 and frame 700 of the MOT17-11 sequence. The occluded person is highlighted by the red dotted ellipse. The distance value is computed by the track and detection of the occluded person in frame 700. The occluded person is not detected from frame 687 to frame 699 and it is set as a lost track.  
	}
	\label{fig:occlusion}
	\vspace{-3mm}
\end{figure*}

Another reason for the IDF1 performance gain of our method is that marginal probability is more robust to occlusion. We show some detailed visualization results of how our approach deals with occlusion and retrieve the lost object. As is shown in the second line of Figure~\ref{fig:occlusion}, our approach can keep the identity unchanged in the case of severe occlusion. We adopt Kalman Filter to filter some impossible matchings (distance is infinity). The distance matrix in Figure~\ref{fig:occlusion} is the input of the Hungarian Algorithm in the first matching. The distance of the occluded person is 0.753 in the distance-based method and the distance is 0.725 in the probability-based method. From the distance distribution figure, we can see that the optimal matching threshold is 0.6 for the distanced-based method and 0.8 for the probability-based method. Only the probability-based method can preserve the identity of the occluded person as 0.725 is smaller than 0.8. Because we consider all possible matchings to compute the marginal probability, the probability of many impossible pairs can be very low (\eg 0) and the distance is very large (\eg 1) after using 1 to minus the probability. After adding the motion distance, the final distance is compressed between 0.8 to 1.2 and thus we can set a relatively high matching threshold (\eg 0.8) and successfully retrieve the lost object with a large distance (\eg 0.7). Also, by comparing 0.725 to 0.753, we can see that the marginal probability can make the distance of correct-matched occluded pairs lower.

\subsection{Ablation studies}
In this section, we compare different methods to compute marginal probability and evaluate different components of our matching strategy. We also evaluate the time-consuming of our method. We adopt Scaled-YOLOv4 \cite{wang2020scaled} as the detector, BoT \cite{luo2019bag} as the Re-ID model and evaluate on MOT17 training sets. We utilize powerful deep learning models and domain-different training data and evaluation data to make our setting as close to real-world applications as possible, which can better reflect the real performance of our method.

\myparagraph{Probability computation. } 
We compare different methods to compute the marginal probability, including softmax, bi-directional softmax, and our marginal inference method. The softmax method is to compute softmax probability between each track and all the detections. Bi-directional softmax is to compute probability between each track and all the detections along with each detection and all the tracks. We average the two probabilities to get the final probability. 

The results are shown in Table~\ref{table:softmax}. We can see that the bi-directional softmax method has the highest MOTA and lowest ID switches while our method has the highest IDF1 score. The softmax-based methods only consider one-to-n matchings and lack global consideration. Our method can approximate all possible matchings and thus has global consideration. Our method has sightly more ID switches than bi-softmax because sometimes the ID will switch 2 times and then turn to be correct in some cases of severe occlusion. In such cases, the IDF1 score is still high and we argue that IDF1 score is more important. 

\begin{table}
\begin{center}
\setlength{\tabcolsep}{10pt}
\begin{tabular}{llll}
\toprule
Method & MOTA$\uparrow$ & IDF1$\uparrow$ & IDs$\downarrow$\\
\midrule
Distance & 71.5 & 74.8 & 499\\
Softmax & 71.6 & 73.7 & 477\\
Bi-softmax & 71.6 & 74.7 & \textbf{405}\\
Marginal (Ours) & \textbf{71.6} & \textbf{75.7} & 449\\
\bottomrule
\end{tabular}
\end{center}
\caption{Comparison of different methods to compute probability. ``Bi-softmax'' is short for bi-directional softmax. ``Marginal'' is short for marginal inference. }
\vspace{-5mm}
\label{table:softmax}
\end{table}

\myparagraph{Matching strategy. }
We evaluate the effect of different components in the matching strategy, including appearance features, sparse probability, Kalman Filter and IoU. As is shown in Table~\ref{table:matching}, appearance and motion cues are complementary. IoU matching often works when appearance features are unreliable (\eg occlusion). Finally, the marginal probability matching further increase the IDF1 score by about 1 point. 

\begin{table}
\begin{center}
\setlength{\tabcolsep}{10pt}
\begin{tabular}{lllllll}
\toprule
A & K & IoU & P & MOTA$\uparrow$ & IDF1$\uparrow$ & IDs$\downarrow$\\
\midrule
$\checkmark$ &  &  &  & 68.8 & 71.9 & 792\\
$\checkmark$ & $\checkmark$ &  &  & 70.1 & 73.6 & 777\\
$\checkmark$ & $\checkmark$ & $\checkmark$ &  & 71.5 & 74.8 & 499\\
$\checkmark$ & $\checkmark$ & $\checkmark$ & $\checkmark$ & \textbf{71.6} & \textbf{75.7} & \textbf{449}\\
\bottomrule
\end{tabular}
\end{center}
\caption{Ablation study of different components in the matching strategy. ``A'' is short for appearance features, ``P'' is short for probability, ``K'' is short for Kalman Filter. }
\vspace{-5mm}
\label{table:matching}
\end{table}

\begin{figure}
	\centering
	\vspace{-3mm}
	\includegraphics[width=2.5in]{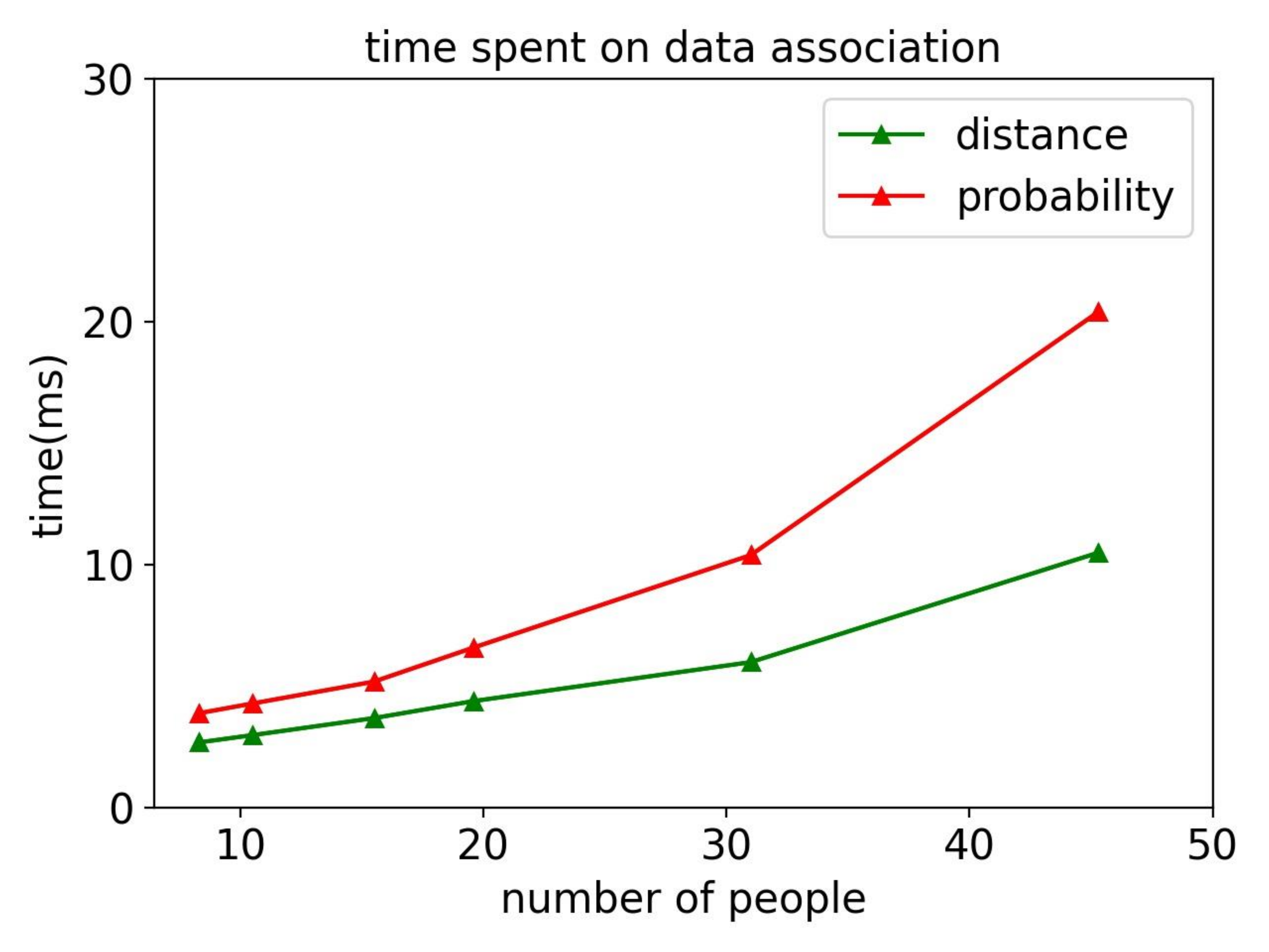}
	\caption{Visualization of the time-consuming of data association. We evaluate two different methods on the training set of MOT17. 
	}
	\label{fig:time}
	\vspace{-3mm}
\end{figure}

\myparagraph{Time-consuming. }
We compute the time-consuming of the linking step using videos with different density (average number of pedestrians per frame). We compare the distance-based matching method to the probability-based matching method. We choose videos with different density from the MOT17 training set. As is shown in Figure~\ref{fig:time}, it takes only a fraction of time (less than 10 ms) to compute the marginal probability.

\begin{figure}
	\centering
	\includegraphics[width=2.5in]{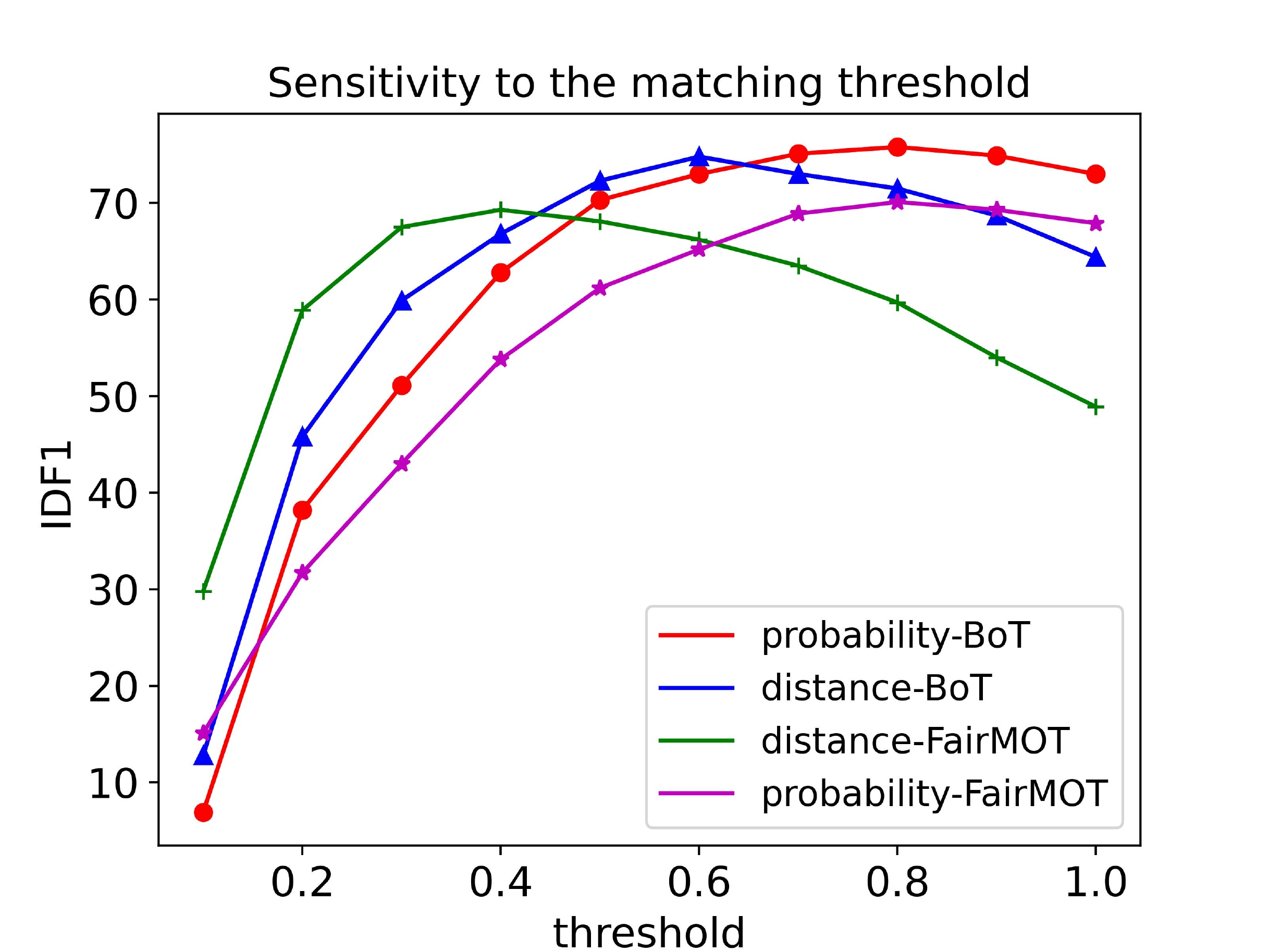}
	\caption{Visualization of how IDF1 score changes with the matching threshold. We evaluate two different methods with distance or probability on the training set of MOT17. 
	}
	\label{fig:threshold}
	\vspace{-3mm}
\end{figure}

\myparagraph{Sensitivity of the matching threshold. }
We set the matching threshold from 0.1 to 1.0 to see how IDF1 score changes with the threshold on the training set of MOT17. We use detection results from Scaled-YOLOv4 and Re-ID features from BoT and FairMOT. From Figure~\ref{fig:threshold} we can see that the probability-based method achieves stable IDF1 when the threshold ranges from 0.7 to 1.0, which is more robust to the distance-based method.

\setlength{\tabcolsep}{5pt}
\begin{table}[!htb]
\footnotesize
\begin{center}
\begin{tabular}{llccccccc}
\toprule
\multicolumn{9}{c}{Public Detection} \\
\midrule
Mode & Method & MOTA$\uparrow$ & IDF1$\uparrow$ & MT$\uparrow$ & ML$\downarrow$ & FP$\downarrow$ & FN$\downarrow$ & IDs$\downarrow$\\
\midrule
Off & MHT\_DAM \cite{kim2015multiple} & 50.7 & 47.2 & 491 & 869 & 22875 & 252889 & 2314 \\
Off & jCC \cite{keuper2018motion} & 51.2 & 54.5 & 493 & 872 & 25937 & 247822 & 1802 \\
Off & FWT \cite{henschel2018fusion} & 51.3 & 47.6 & 505 & 830 & 24101 & 247921 & 2648 \\
Off & eHAF \cite{sheng2018heterogeneous} & 51.8 & 54.7 & 551 & 893 & 33212 & 236772 & 1834 \\
Off & TT \cite{zhang2020long} & 54.9 & 63.1 & 575 & 897 & 20236 & 233295 & 1088 \\
Off & MPNTrack \cite{braso2020learning} & 58.8 & 61.7 & \textbf{679} & \textbf{788} & 17413 & 213594 & \textbf{1185} \\
Off & Lif\_T \cite{hornakova2020lifted} & \textbf{60.5} & \textbf{65.6} & 637 & 791 & \textbf{14966} & \textbf{206619} & 1189 \\
\midrule
On & MOTDT \cite{chen2018real} & 50.9 & 52.7 & 413 & 841 & 24069 & 250768 & 2474 \\
On & FAMNet \cite{chu2019famnet} & 52.0 & 48.7 & 450 & 787 & 14138 & 253616 & 3072 \\
On & DeepMOT \cite{xu2020train} & 53.7 & 53.8 & 458 & 861 & 11731 & 247447 & 1947 \\
On & Tracktor++v2 \cite{bergmann2019tracking} & 56.3 & 55.1 & 498 & 831 & \textbf{8866} & 235449 & 1987 \\
On & CenterTrack \cite{zhou2020tracking} & 61.5 & 59.6 & 621 & 752 & 14076 & 200672 & 2583 \\
On & MTracker (Ours) & \textbf{62.1} & \textbf{65.0} & \textbf{657} & \textbf{730} & 24052 & \textbf{188264} & \textbf{1768} \\
\midrule
\multicolumn{9}{c}{Private Detection} \\
\midrule
Mode & Method & MOTA$\uparrow$ & IDF1$\uparrow$ & MT$\uparrow$ & ML$\downarrow$ & FP$\downarrow$ & FN$\downarrow$ & IDs$\downarrow$\\
\midrule
On & TubeTK \cite{pang2020tubetk} & 63.0 & 58.6 & 735 & 468 & 27060 & 177483 & 4137 \\
On & CTracker \cite{peng2020chained} & 66.6 & 57.4 & 759 & 570 & 22284 & 160491 & 5529 \\
On & CenterTrack \cite{zhou2020tracking} & 67.8 & 64.7 & 816 & 579 & \textbf{18489} & 160332 & \textbf{3039} \\
On & FairMOT \cite{zhang2021fairmot} & 73.7 & 72.3 & 1017 & 408 & 27507 & 117477 & 3303 \\
On & PermaTrackPr \cite{tokmakov2021learning} & 73.8 & 68.9 & 1032 & 405 & 28998 & 115104 & 3699 \\
On & TransTrack \cite{sun2020transtrack} & 75.2 & 63.5 & 1302 & \textbf{240} & 50157 & 86442 & 3603 \\
On & CorrTracker \cite{wang2021multiple} & 76.5 & 73.6 & 1122 & 300 & 29808 & 99510 & 3369 \\
On & MTracker (Ours) & \textbf{77.3} & \textbf{75.9} & \textbf{1314} & 276 & 45030 & \textbf{79716} & 3255 \\
\bottomrule

\end{tabular}
\end{center}
\caption{Comparison of the state-of-the-art methods on MOT17 test sets. We report results under both public detection and private detection protocols. }
\vspace{-8mm}
\label{table:sota17}
\end{table}

\setlength{\tabcolsep}{5pt}
\begin{table}[!htb]
\footnotesize
\begin{center}
\begin{tabular}{llccccccc}
\toprule
\multicolumn{9}{c}{Public Detection} \\
\midrule
Mode & Method & MOTA$\uparrow$ & IDF1$\uparrow$ & MT$\uparrow$ & ML$\downarrow$ & FP$\downarrow$ & FN$\downarrow$ & IDs$\downarrow$\\
\midrule
Off & IOU19 \cite{bochinski2017high}* & 35.8 & 25.7 & 126 & 389 & 24427 & 319696 & 15676 \\
Off & V-IOU \cite{bochinski2018extending}* & 46.7 & 46.0 & 288 & 306 & 33776 & 261964 & 2589 \\
Off & MPNTrack \cite{braso2020learning} & \textbf{57.6} & \textbf{59.1} & \textbf{474} & \textbf{279} & \textbf{16953} & \textbf{201384} & \textbf{1210} \\
\midrule
On & SORT20 \cite{bewley2016simple} & 42.7 & 45.1 & 208 & 326 & 27521 & 264694 & 4470 \\
On & TAMA \cite{yoon2020online}* & 47.6 & 48.7 & 342 & \textbf{297} & 38194 & 252934 & 2437 \\
On & Tracktor++ \cite{bergmann2019tracking}* & 51.3 & 47.6 & 313 & 326 & 16263 & 253680 & 2584 \\
On & Tracktor++v2 \cite{bergmann2019tracking} & 52.6 & 52.7 & 365 & 331 & \textbf{6930} & 236680 & 1648 \\
On & MTracker (Ours) & \textbf{55.6} & \textbf{65.0} & \textbf{444} & 388 & 12297 & \textbf{216986} & \textbf{480} \\
\midrule
\multicolumn{9}{c}{Private Detection} \\
\midrule
Mode & Method & MOTA$\uparrow$ & IDF1$\uparrow$ & MT$\uparrow$ & ML$\downarrow$ & FP$\downarrow$ & FN$\downarrow$ & IDs$\downarrow$\\
\midrule
On & MLT \cite{zhang2020multiplex} & 48.9 & 54.6 & 384 & 274 & 45660 & 216803 & \textbf{2187} \\
On & FairMOT \cite{zhang2021fairmot} & 61.8 & 67.3 & \textbf{855} & \textbf{94} & 103440 & \textbf{88901} & 5243 \\
On & TransTrack \cite{sun2020transtrack} & 65.0 & 59.4 & 622 & 167 & 27197 & 150197 & 3608 \\
On & CorrTracker \cite{wang2021multiple} & 65.2 & 69.1 & - & - & 79429 & 95855 & 5183 \\
On & MTracker (Ours) & \textbf{66.3} & \textbf{67.7} & 707 & 146 & \textbf{41538} & 130072 & 2715 \\
\bottomrule

\end{tabular}
\end{center}
\caption{Comparison of the state-of-the-art methods on MOT20 test sets. We report results under both public detection and private detection protocols. The methods denoted by * are the ones reported on CVPR2019 Challenge in which the videos and ground-truth are almost the same as MOT20. }
\vspace{-5mm}
\label{table:sota20}
\end{table}

\subsection{Benchmark evaluation}
We compare our Marginal Inference Tracker (MTracker) with the state-of-the-art methods on the test sets of MOT17 and MOT20 under both public detection and private detection protocols. We list the results of both online methods and offline methods for completeness. We only compare directly to the online methods for fairness. For public detection results, we adopt the one-shot tracker FairMOT \cite{zhang2021fairmot} to jointly perform detection and Re-ID and follow CenterTrack \cite{zhou2020tracking} to use public detections to filter the tracklets with a more strict IoU distance. For private detection results, we adopt a more powerful detector Scaled-Yolov4 \cite{wang2020scaled} and Re-ID model BoT \cite{luo2019bag}.

Table~\ref{table:sota17} and Table~\ref{table:sota20} show our results on the test sets of MOT17 and MOT20. For public detection results, MTracker achieves high IDF1 score and low ID switches and outperforms the state-of-the-art methods by a large margin. On MOT17 test sets, the IDF1 score of MTracker is 5.4 points higher than CenterTrack and the ID switches are reduced by 30\%. On MOT20 test sets, the IDF1 score of MTracker is 12.3 points higher than Tractor++v2 \cite{bergmann2019tracking} and the ID switches are reduced by 70\%. The high IDF1 score and low ID switches indicate that our method has strong identity preservation ability, which reveals the advantages of the marginal probability. For the private detection results, we use the same training data as FairMOT and substantially outperforms it on both MOTA and IDF1 score.

\section{Conclusion}
We present an efficient and robust data association method for multi-object tracking by marginal inference. The obtained marginal probability can be regarded as ``normalized distance'' and is significantly more stable than the distances based on Re-ID features. Our probability-based data association method has several advantages over the classic distance-based one. First, we can use a single threshold for all videos thanks to the stable probability distribution. Second, we empirically find that marginal probability is more robust to occlusion. Third, our approach is general and can be applied to the existing state-of-the-art trackers \cite{zhang2021fairmot,wojke2017simple} easily. We hope our work can benefit real applications where data distribution always varies significantly. 

\myparagraph{Acknowledgement }
This work was in part supported by NSFC (No. 61733007 and No. 61876212) and MSRA Collaborative Research Fund.

\clearpage
%
%
\bibliographystyle{splncs04}
\bibliography{egbib}
\end{document}